\documentclass[10pt,twocolumn,letterpaper,nameinlink,capitalize]{article}

\usepackage[final]{iccv}      

\usepackage{multirow}
\usepackage{makecell}
\usepackage[ruled,vlined]{algorithm2e}
\usepackage[font=small]{caption}
\usepackage{bm}
\usepackage{wrapfig}
\usepackage{colortbl} 
\usepackage{floatrow}
\usepackage{color}
\usepackage{pifont}
\newcommand{\cmark}{\ding{51}} 
\newcommand{\xmark}{\ding{55}} 
\usepackage{graphicx}

\definecolor{red}{rgb}{0.95,0.4,0.4}
\definecolor{blue}{rgb}{0.4,0.4,0.95}
\definecolor{darkblue}{rgb}{0,0,0.8}
\definecolor{darkred}{rgb}{0.8,0,0}
\definecolor{darkblue}{rgb}{0,0.5,0}
\definecolor{grey}{rgb}{0.6,0.6,0.6}
\definecolor{col1}{RGB}{232, 161, 148}
\definecolor{col2}{RGB}{148, 187, 232}

\usepackage{amsmath}
\usepackage{makecell}
\usepackage[utf8]{inputenc}
\usepackage{graphicx}
\usepackage{subcaption}  
\DeclareUnicodeCharacter{2713}{\checkmark}

\definecolor{iccvblue}{rgb}{0.21,0.49,0.74}
\usepackage[pagebackref,breaklinks,colorlinks,allcolors=iccvblue]{hyperref}

\def\paperID{} 
\def\confName{}
\def\confYear{}

\title{Region-aware Depth Scale Adaptation with Sparse Measurements}

\author{Rizhao Fan\\
\and
Tianfang Ma
\and
Zhigen Li\\
\\
\and
Ning An
\and
Jian Cheng
\\
}

\begin{document}
\maketitle
\begin{abstract}

In recent years, the emergence of foundation models for depth prediction has led to remarkable progress, particularly in zero-shot monocular depth estimation. These models generate impressive depth predictions; however, their outputs are often in relative scale rather than metric scale. This limitation poses challenges for direct deployment in real-world applications. To address this, several scale adaptation methods have been proposed to enable foundation models to produce metric depth. However, these methods are typically costly, as they require additional training on new domains and datasets. Moreover, fine-tuning these models often compromises their original generalization capabilities, limiting their adaptability across diverse scenes. In this paper, we introduce a non-learning-based approach that leverages sparse depth measurements to adapt the relative-scale predictions of foundation models into metric-scale depth. Our method requires neither retraining nor fine-tuning, thereby preserving the strong generalization ability of the original foundation models while enabling them to produce metric depth. Experimental results demonstrate the effectiveness of our approach, highlighting its potential to bridge the gap between relative and metric depth without incurring additional computational costs or sacrificing generalization ability.

\end{abstract}    
\section{Introduction}
\label{sec:intro}

\begin{figure}[t] 
    \centering
    \begin{subfigure}[b]{0.495\linewidth}
        \centering
        \includegraphics[width=\linewidth]{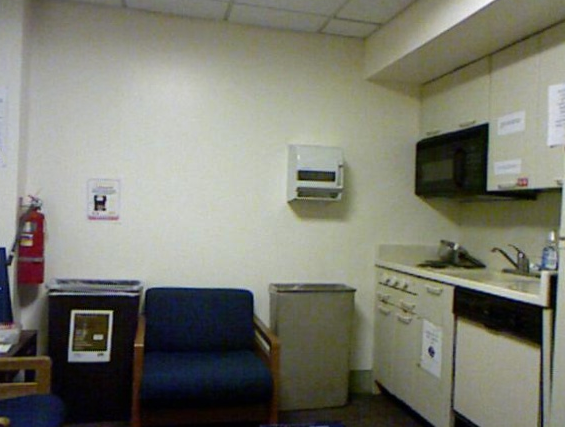}  
        \caption*{(a)}
        \label{fig:rgb}
    \end{subfigure}
    \hfill
    \begin{subfigure}[b]{0.495\linewidth}
        \centering
        \includegraphics[width=\linewidth]{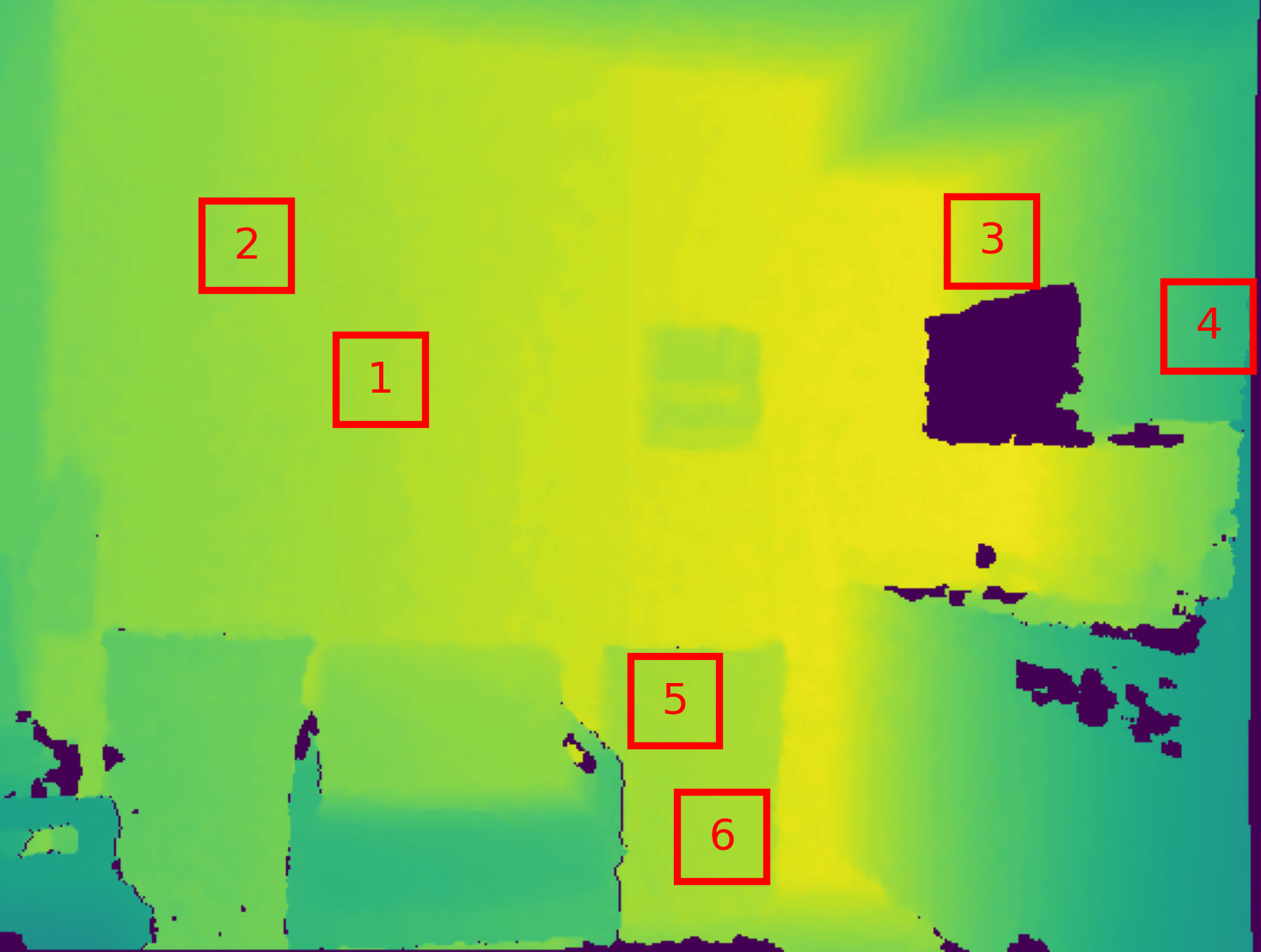} 
        \caption*{(b)}
        \label{fig:depth_gt}
    \end{subfigure}
    \vspace{5pt}  
    \begin{subfigure}[b]{0.495\linewidth}
        \centering
        \includegraphics[width=\linewidth]{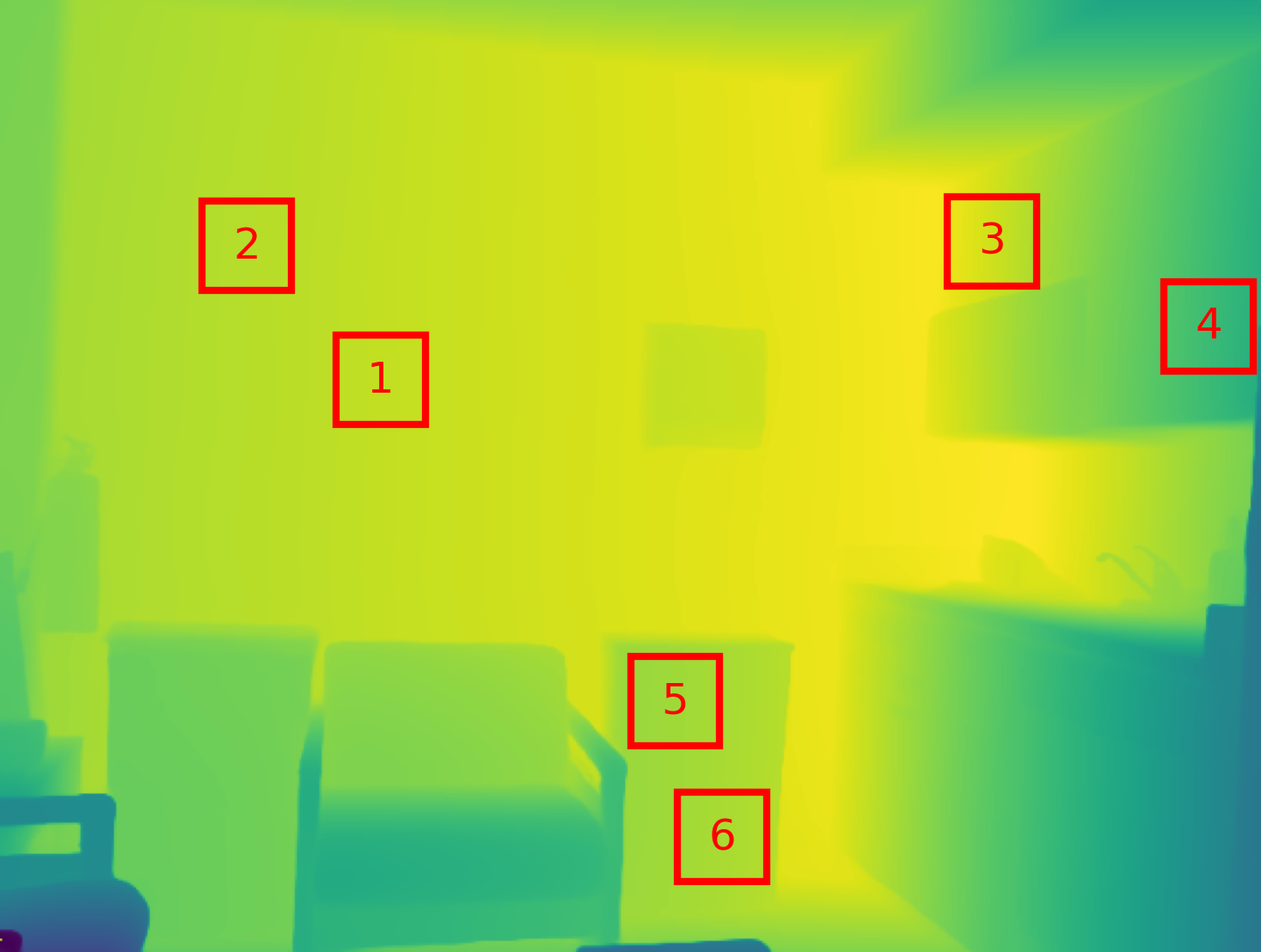} 
        \caption*{(c)}
        \label{fig:depth_pred}
    \end{subfigure}
    \hfill
    \begin{subfigure}[b]{0.45\linewidth}
        \centering
\begin{small}  
    \begin{tabular}{c|c|c}
        \toprule
        \textbf{Patch} & \textbf{Scale (s)} & \textbf{Shift (t)} \\
        \midrule
Patch 1  & 2.68      &-14.13 \\
Patch 2  & 3.03      &-17.38 \\
Patch 3  & 1.45      & -2.00 \\
Patch 4  & 1.74      & -3.52 \\
Patch 5  & 0.78      &  5.00 \\
Patch 6  & 0.61      &  6.52 \\
        \bottomrule
    \end{tabular}
\end{small}
        \caption*{(d)}
        \label{fig:table}
    \end{subfigure}

    \caption{Analysis of Scale and Shift factor in MDE for a selected scene.
(a) The RGB input image.  
(b) The ground-truth depth map, where six selected patches are highlighted.  
(c) The Depth Anything v2 predicted depth map, with patches at the same locations as (b).  
(d) Computed scale and shift factors by comparing (b) with (c) for each selected patch.}
    \label{fig:depth_comparison}
\end{figure}
\vspace{1cm}

Depth estimation, which aims to recover the 3D structure of scenes and serves as an important task in various computer vision applications, including robotic navigation, autonomous driving, and augmented reality. With the advent of deep learning, significant progress has been made in monocular depth estimation (MDE) tasks \cite{yuan2022new, aich2021bidirectional, lee2021patch, lee2019big}.
Recent advancements in vision foundation models have introduced new paradigms for visual perception tasks. 
Segment Anything \cite{kirillov2023segment, ravi2024sam}, along with subsequent works following its approach \cite{chen2024ma, chen2023sam, zhang2023faster, wu2023medical, zhang2023personalize}, 
has demonstrated remarkable versatility in image segmentation by accurately delineating objects across diverse domains with minimal input.
Building on this foundation model philosophy, Depth Anything \cite{depth_anything, depth_anything_v2} and related approaches \cite{bochkovskii2024depth, lin2024promptda} have achieved high-quality monocular depth predictions by leveraging large-scale diverse training data and advanced architectures. 
While these foundation models demonstrate impressive generalization capabilities, depth estimation models that rely solely on image inputs face challenges such as scale ambiguity, making them difficult to apply directly in real-world applications.

Several efforts \cite{bhat2023zoedepth, zeng2024rsa, lin2024promptda, bochkovskii2024depth, park2024depth} have been made to adapt existing depth foundation models for metric depth estimation tasks. These approaches often require retraining and fine-tuning on new datasets or depend on additional prompts. ZoeDepth \cite{bhat2023zoedepth} employs a bin adjustment head to adapt relative depth pre-training for metric fine-tuning in new domains. \cite{park2024depth} introduced a depth prompt module designed to work with foundation models for MDE. However, these methods are not only computationally expensive but may also compromise the generalization ability of the original depth estimation foundation models. 
A recent work \cite{marsal2024foundation} proposed a non-learning-based approach converting relative depth outputs into metric depth by introducing 3D points provided by low-cost sensors or techniques.

However, these above methods have significant limitations. 
All of the above methods rely on a global scaling parameter to recover metric depth. In the field of MDE, perspective projection causes depth ambiguity, where a single 2D image can correspond to many possible 3D scenes. As a result, applying a global scaling factor to the entire depth map introduces significant errors in depth recovery. As illustrated in our analysis in Figure \ref{fig:depth_comparison}), a comparison between the relative depth map predicted by Depth Anything v2 \cite{depth_anything_v2} and the ground truth data reveals substantial differences in scaling parameters between different objects. In contrast, regions from the same object, exhibit similar scaling parameters. This variability demonstrates that a single global scaling factor fails to accurately handle the diverse depth relationships in complex scenes, making it an ineffective solution.

To address this challenge, we propose leveraging sparse depth measurements to adapt foundation model outputs from relative to metric depth. Unlike existing methods that apply a global scaling factor \cite{marsal2024foundation, zeng2024rsa, godard2019digging, ranftl2020towards}, we assign distinct scaling parameters to different regions within the scene. A key step is segmenting the scene into meaningful regions, whereas window-based partitioning \cite{liu2021swin, dong2022cswin, fan2023contrastive} may mix pixels from different objects, and superpixel-based methods \cite{Fan2018, Dai2022, Teutscher2021} risk over-segmentation. Instead, we leverage foundation models such as Segment Anything \cite{kirillov2023segment} and OneFormer \cite{jain2023oneformer} to segment scenes based on color, texture, shape, or brightness, aligning with entire objects or meaningful parts. 
We introduce a novel method to convert the scale-ambiguous depth predictions of foundation models into metric depth using sparse depth measurements. This approach assigns a unique scaling factor to each segmented region, enabling region-aware adjustments from relative to metric depth. Sparse depth measurements are utilized to compute these scaling factors, ensuring accurate calibration. By introducing region-aware scaling, our method achieves finer granularity and higher accuracy than global scaling techniques while eliminating the need for costly retraining and fine-tuning on target datasets. Extensive experiments on standard depth estimation benchmarks validate its effectiveness across diverse scenarios, offering a practical and adaptable solution to mitigate scale ambiguity in MDE.

Our key contributions are as follows:  
\begin{itemize}
\item 
We provide an in-depth analysis of existing depth scale recovery methods and highlight their limitations. First, a single global scaling factor is inadequate, as different objects exhibit distinct scale factors. Second, current methods treat the depth map as a collection of independent numerical values and fit a simple scale-shift transformation, failing to capture its nature. Instead, the depth map should be modeled as a composition of multiple structured surfaces to ensure accurate metric depth recovery.

\item We propose a region-aware depth scaling adaptation method for MDE foundation models. Our approach segments the scene into regions and assigns distinct scaling factors, obtained through sparse depth measurements, enabling more precise metric depth recovery.

\item We conduct comprehensive experiments across multiple datasets, demonstrating the effectiveness of our method. Our approach consistently outperforms global scaling strategies, achieving higher accuracy in metric depth estimation.  
\end{itemize}

\section{Related Work}
\label{sec:related_work}
In this section, we review self-supervised depth estimation approaches relevant to our work.
(1) monocular depth estimation, 
(2) vision foundation models,
(3) depth scale adaptation methods for depth estimation models.

\subsection{Monocular Depth Estimation}

Monocular Depth Estimation (MDE) is a core task in computer vision, playing a important role in transforming 2D images into 3D scene geometry \cite{fu2018deep, yuan2022new, liu2023va, bhat2021adabins, wu2022toward}. 
The advancement of MDE has been significantly driven by deep learning-based methods \cite{geiger2013vision, silberman2012indoor}.
Eigen et al. \cite{eigen2014depth} started a breakthrough in MDE by developing a multi-scale fusion network. 
Since then, numerous works \cite{bhat2021adabins, yin2019enforcing, yin2021learning, li2022binsformer} have been proposed to continuously improve MDE prediciton accuracy. 
AdaBins \cite{bhat2021adabins} partitions depth ranges into adaptive bins, estimating final depth values as linear combinations of the bin centers. 
Nddepth \cite{shao2023nddepth} incorporates geometric priors through a physics-driven deep learning approach. 
NeWCRFs \cite{yuan2022new} utilizes neural window fully-connected CRFs to optimize energy computation. 
UniDepth \cite{piccinelli2024unidepth} enables the reconstruction of metric 3D scenes from single images across different domains. 
DCDepth \cite{wang2024dcdepth} formulates MDE as a progressive regression task in the discrete cosine domain, further enhancing depth estimation performance.
MiDaS \cite{ranftl2020towards}  introduced a scale-invariant monocular depth estimation approach by training on mixed multi-source datasets and designing a scale-invariant loss function, achieving strong zero-shot cross-dataset generalization.
LeReS \cite{yin2021learning} propose a scale-invariant depth estimation framework with a novel depth normalization technique to handle diverse datasets.

\subsection{Vision Foundation Models}

{The Foundation Models are reshaping computer vision tasks.}
Segment Anything (SAM) \cite{kirillov2023segment} is a groundbreaking foundation model in computer vision, achieving high-precision, class-agnostic segmentation with strong zero-shot capabilities. It combines a ViT-based image encoder, a lightweight mask decoder, and a flexible prompt encoder supporting points, boxes, masks, and text inputs. SAM advances interactive segmentation and demonstrates exceptional adaptability across diverse tasks, significantly expanding the scope of computer vision research.
Recent works \cite{chen2024ma, chen2023sam, zhang2023faster, zhang2023personalize, wu2023medical} have been dedicated to exploring various variants of SAM to further enhance performance.

Following the design philosophy of foundation models, the Depth Anything series \cite{depth_anything, depth_anything_v2} was introduced. This approach proposes a robust monocular depth estimation framework that leverages millions of training samples to develop more powerful depth estimators, achieving remarkable zero-shot depth accuracy across diverse scenes.
Depth Pro \cite{bochkovskii2024depth} is a foundation model specifically designed for zero-shot metric monocular depth estimation. It is capable of generating high-resolution metric depth maps with absolute scale, making it a strong contender in this domain.
Several methods \cite{he2024lotus, ke2024repurposing, gui2024depthfm} utilize diffusion-based visual foundation models to synthesize high-quality relative depth maps, further advancing the field of depth estimation.

\subsection{Scale Adaptation for Monocular Depth Estimation}
To transform the predicted relative depth into metric depth, some studies have made significant attempts. 
\cite{park2024depth} proposed a sparse depth prompt and integrate it with foundation models for monocular depth estimation to generate absolute-scale depth maps. 
ZoeDepth \cite{bhat2023zoedepth} and Depth Anything \cite{depth_anything} utilize a metric bins module within the decoder to compute per-pixel depth bin centers, which are then linearly combined to produce metric depth. MfH \cite{zhao2024metric} propagates metric information from annotated human figures to other parts of the scene, thereby generating metric depth estimates for the original input images. Monodepth2 \cite{godard2019digging} utilizes a per-image median ground truth scaling approach when measuring errors. 
MiDas \cite{ranftl2020towards} aligns predictions and ground truth in scale and shift for each image in inverse-depth space based on the least-square criterion when measuring errors. 
DistDepth \cite{wu2022toward} integrates metric scale into a scale-agnostic depth network by leveraging left-right stereo consistency. 
RSA \cite{zeng2024rsa} generates scale using text to transfer relative depth to metric depth across domains and does not require ground truth during test time. 
ScaleDepth \cite{zhu2024scaledepth} decomposes metric depth estimation into two dedicated modules: one for relative depth estimation and another for scale estimation, and it can also leverage textual descriptions of the scene to guide the supervision process. 
\cite{marsal2024foundation} proposed estimating the scale factor from low-cost sensors to enhance the predicted relative depth results of foundation models.

The MDE models exhibit certain limitations. 
Some depth estimation models suffer from poor generalization and lack zero-shot capabilities, 
while 
others demonstrate strong generalization and zero-shot abilities but can only produce inverse depth relative results. 
To achieve depth adaptation from relative to metric, some approaches require retraining on new datasets, while others rely on additional prompts. Moreover, these methods overlook the intrinsic nature of depth maps, treating depth data merely as a set of values for linear transformation, while ignoring the fundamental fact that depth maps inherently represent planar structures.

\section{Methodology}

\begin{figure*}
  \centering
\includegraphics[height=55mm]{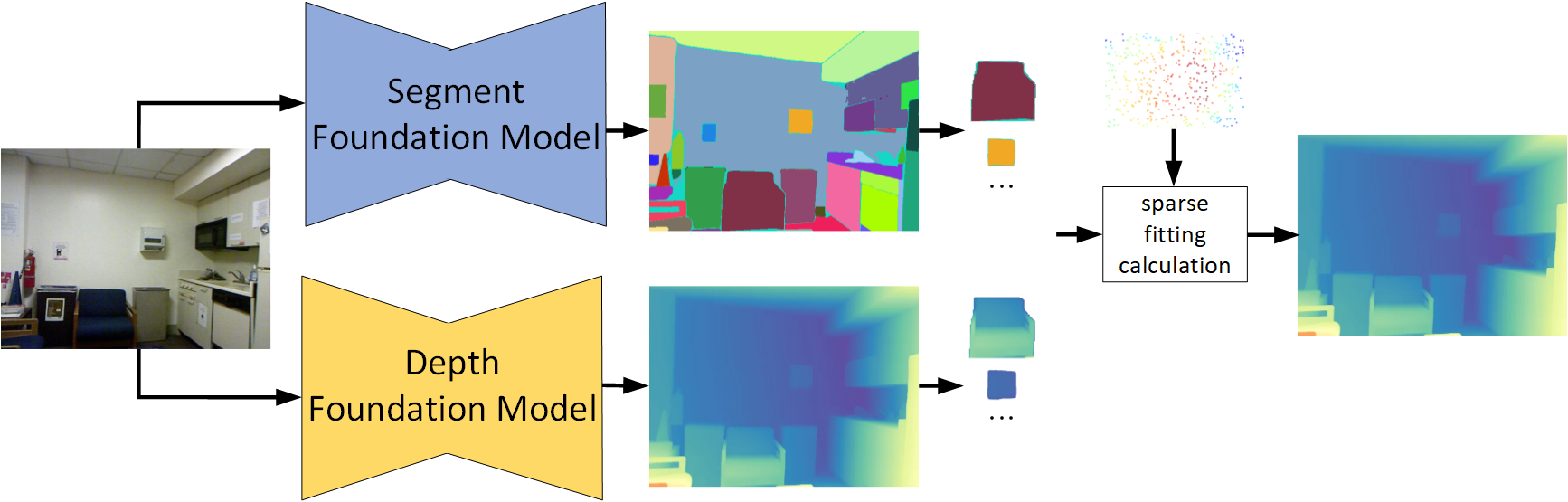}
  \caption{\textbf{Our proposed framework.} 
The input image is processed by a Segmentation Foundation Model and a MDE Foundation Model, generating a segmentation map $M$ and a relative depth prediction $D$.  
$M$, $D$ are divided into multiple small regions.  
Within each region, sparse fitting calculations are applied to obtain a metric-scaled depth map.
Finally, the metric-scaled depth maps from all regions are merged to produce the final depth result.
}
  \label{fig:overall_design}
\end{figure*}

Given an RGB image $ I \in \mathbb{R}^{H \times W \times 3} $, monocular depth estimation foundation models $ \mathcal{F}_d $ 
predicts an inverse-depth map $ d \in \mathbb{R}^{H \times W} $,
a segmentation model $ \mathcal{F}_s $ 
generates a segmentation mask $ M \in \mathbb{R}^{H \times W} $.
Since $d$ is inversely related to depth, we first transform $ d $ into a relative depth map $ D \in \mathbb{R}^{H \times W} $. However, this estimation result $ D $ 
has an ambiguous scale. A common approach is to obtain a metric depth map $ D_m $ using an transformation with scale and shift:

\begin{equation}
D_m = \alpha D + \beta, \quad (\alpha, \beta) \in \mathbb{R}^2
\end{equation}

where $ \alpha $ and $ \beta $ are the global scale and shift parameters, respectively. 
Previous studies \cite{ranftl2020towards, zhu2024scaledepth, marsal2024foundation, zeng2024rsa} have explored different strategies for estimating these parameters. Specifically, \cite{marsal2024foundation} utilizes low-cost sensors to obtain $ \alpha $ and $ \beta $, while \cite{zeng2024rsa} infers scale from language. 
However, 
using \textbf{a single global scale} is insufficient for accurate metric depth recovery.

\subsection{Limitations of Global Scale and Shift Estimation}


To analyze this limitation, we evaluate the depth prediction of $ \mathcal{F}_d$ on a sample from the NYU Depth v2 dataset and compare it with the ground truth depth $ D_{\text{GT}} $. 
We first apply an {affine-invariant transformation} to the predicted relative depth maps:

\begin{align}
D' &= \frac{D - t(D)}{s(D)},
\end{align}

where $ t(D) $ and $ s(D) $ are used to ensure zero translation and unit scale:

\begin{align}
t(D) = \text{median}(D),  s(D) = \frac{1}{HW} \sum_{i,j} |D_{i,j} - t(D)|\!\!\!\!
\end{align}

In Figure~\ref{fig:depth_comparison}, we illustrate the scale and shift parameters between $ D_{\text{GT}} $ and $ D' $ across multiple sub-regions within the image. 
It is evident that these parameters are \textbf{not uniform across the entire scene}. Specifically, different objects exhibit distinct scale and shift factors—for example, patches 1 (from the wall), patches 3 (from the wall cabinet),  and patches 5 (from the rubbish bin) demonstrate noticeable differences. 
In contrast, patches belonging to the same object—such as regions 1 and 2 from the wall, and regions 3 and 4 from the cabinet—tend to exhibit more consistent values.
These observations suggest that a \textbf{single global scaling factor} is insufficient for accurately recovering metric depth across the entire image.  

Instead, a more effective approach is to assign \textbf{region-aware} scale and shift factors based on the segmentation mask $ M $, rather than relying on a global scaling factor. In the following sections, we introduce a method that leverages region-aware transformations to enhance the accuracy of relative-to-metric depth conversion.

\subsection{Region-aware Depth Scaling with Sparse Measurements}

Given a relative depth map $ D \in \mathbb{R}^{H \times W} $ and a segmentation mask $ M \in \mathbb{R}^{H \times W} $, where $ M $ consists of $ i $ regions, we treat $ D $ as a collection of individual regions $\ D_0, D_1, \dots, D_i $ based on the segmentation mask $M_0, M_1, \dots, M_i $. As previously discussed, each of these regions $ D_i $ should have its own independent scaling factor and shift. This allows us to better handle variations in depth across different image regions and improves the accuracy of depth estimation.

We use a strategy similar to depth completion, sparse depth measurements are utilized to guide the transformation of the relative depth map $ D $ into a metric depth map. However, unlike depth completion, where sparse depth measurements are used to generate a dense depth map, our goal here is to estimate the scaling $\alpha_i$ and shift factor $\beta_i$ for each region $ D_i $. These parameters are then used to map the relative depth map into the metric space.

The sparse depth map $ D_s \in \mathbb{R}^{H \times W} $ contains $ N $ sparse depth measurements, each synchronized with the image $ I \in \mathbb{R}^{H \times W \times 3} $.
When sparse depth measurements are available in the corresponding area of $ D_i $, we use the sparse depth measurements in the region to rescale $ D_i $ through linear regression. This operation ensures that the scale and shift factor for each region $ D_i $ are independently computed, yielding a more accurate and region-specific metric depth map.
When there are no sparse depth measurements, or the measurements are not enough to compute the scaling factor, in the corresponding region $ D_i $, we expand the region $ D_i $ by incorporating neighboring regions $ D_j $
(the region defined by the neighboring segmentation mask $ M_j $ of $ M_i $). This results in a larger region, denoted as $ D_i^+ $, which includes both $ D_i $ and the neighboring regions $ D_j $. 
We then apply the same rescaling linear regression to the expanded region $ D_i^+ $, using the available sparse depth measurements within this enlarged region. If $ D_i^+ $ still does not satisfy the required number of linear regression computation, we continue expanding the region by incorporating additional neighboring regions $ D_k $, until enough sparse depth measurements are available for rescaling. This process is described in Algorithm \ref{alg:region_depth_scaling}.
Through this approach, each $D_i$ is mapped to the metric depth space using its corresponding $\alpha_i$ and $\beta_i$. These are then combined to produce the final metric depth result $D_m$.

\begin{algorithm}[t]
\small
\caption{Region-aware Depth Scaling with Sparse Measurements}
\label{alg:region_depth_scaling}
\KwIn{Relative depth map $ D ~ (D_0, D_1, \dots, D_i)  $, \\
segmentation mask $ M ~ (M_0, M_1, \dots, M_i) $, \\
sparse depth map $ D_s $.}
\KwOut{Metric depth map $ D_{\text{m}}  $.}

Apply affine-invariant transformation to the relative depth map $ D $, \: ~
$ D' = \frac{D - \mu_D}{\sigma_D} $\;

\For{each region $ M_i $ in $ M $}{
    \If{Sparse depth measurements meet the requirements in $ M_i $}{
        Rescale $ D_i $ using linear regression based on the sparse measurements in $ M_i $\;
        $ D_i' = \alpha_i D_i + \beta_i $\;
        Compute scaling factor $ \alpha_i $ and shift factor $ \beta_i $ for $ M_i $\;
    }
    \Else{
        \While{Sparse measurements do not meet the requirements in $ M_i^+ $}{
            Expand $ M_i^+ $ by incorporating neighboring regions $ M_j $\;
            Rescale $ D_i^+ $ using linear regression with the available sparse measurements in $ M_i^+ $\;
            $ D_i^+ = \alpha_{i^+} D_i^+ + \beta_{i^+} $\;
            Compute scaling factor $ \alpha_{i^+} $ and shift factor $ \beta_{i^+} $ for $ M_i^+ $\;
        }
    }
    \textbf{Step 3: Combine the region-specific depth maps}\;
    $ D_{\text{m}} = \bigcup_{i=0}^{n} \left( \alpha_i D_i + \beta_i \right) $\;
}
\end{algorithm}

\subsection{Are Scale and Shift Enough?}

In 2D space, linear regression can be used to fit any two straight lines, and the relationship $ D_m = \alpha D + \beta $ holds when both $ D_m $ and $ D $ are straight lines. However, this does not apply to our problem, as $ D_m $ and $ D $ represent surfaces in 3D space rather than 2D lines.  
More specifically, a depth map is composed of planes in 3D space. Therefore, linear regression is not an ideal method for fitting in this task.

As a result, we shift our approach to a surface fitting method based on least squares, which is better suited for fitting surfaces. This approach allows us to compute the surface parameters by using sparse depth measurements and the corresponding points on the relative depth map.

In this context, the sparse depth measurements, $z_1 $, are a set of discrete points, denoted as:

\vspace{4pt}
$z_1 = \{ (x_1, y_1, z_{1,1}), (x_2, y_2, z_{1,2}), \ldots, (x_n, y_n, z_{1,n}) \}$
\vspace{4pt}

while the relative depth values, $z_2 $, form a dense representation that approximates a continuous surface. Specifically, $z_2 $ can be locally approximated by a plane equation:

\vspace{4pt}
$ z_2 \approx m \cdot x + n \cdot y + l $
\vspace{4pt}

where $z_2 $ represents the relative depth values, and $x, y $ are the corresponding spatial coordinates. This equation illustrates how the relative depth map approximates a plane in 3D space.

To establish a relationship between the sparse depth measurements and the relative depth map, we perform least squares surface fitting using only the valid sparse depth points. Specifically, for the points where $z_1 $ is available, we assume the following relationship:

\begin{align}
z_{1,i} = \alpha \cdot z_{2,i} + \beta \cdot x_i + \gamma \cdot y_i + \delta,  \forall (x_i, y_i, z_{1,i}) \in z_1
\end{align}

where, $z_{1,i} $ are the sparse depth measurements, and $z_{2,i} $ are the corresponding relative depth values at the same coordinates, 
$\alpha $ is scale factor,
$\beta $ and $\gamma$ are slope coefficient factors,
$\delta $ is shift factor.
$\alpha, \beta, \gamma, \delta $ are estimated by fitting this equation only at the sparse depth measurement locations.

Once the fitting is completed, we obtain the optimal factor sets ($ \alpha, \beta, \gamma, \delta $) for all the regions, which are then applied to the dense relative depth map to accurately rescale and shift it, yielding the final metric depth.

\begin{table*}[ht]
    \centering
    \footnotesize
    \begin{tabular}{l|c|c|ccc|ccc}
    \toprule
    \textbf{Models} & \textbf{Scaling} & \textbf{Region-aware} & Abs Rel $\downarrow$ & RMSE $\downarrow$ & $\log _{10} \downarrow$ & $\delta<1.25 \uparrow$ & $\delta<1.25^{2} \uparrow$ & $\delta<1.25^{3} \uparrow$ \\ 
    \midrule 
    ZoeDepth &  Image  & \xmark & 0.077 & 0.282 & 0.033 & 0.951 & 0.994 & 0.999 \\
    \midrule 
    DistDepth & DA          & \xmark &0.289  & 1.077 & -     & 0.706 & 0.934 & -     \\
    DistDepth & DA,Median   & \xmark &0.158  & 0.548 & -     & 0.791 & 0.942 & 0.985 \\
    \hline 
    ZeroDepth & DA        & \xmark & 0.100 & 0.380 & -     & 0.901 & 0.961 & -      \\
    ZeroDepth & DA,Median & \xmark & 0.081 & 0.338 & -     & 0.926 & 0.986 & -      \\
    \hline
    \multirow{13}*{MiDas}
    & Global  & \xmark & 0.183 & 0.600 & 0.078 & 0.689 & 0.949 & 0.992  \\
    & Image   & \xmark & 0.175 & 0.563 & 0.072 & 0.729 & 0.958 & 0.994  \\
    & RSA     & \xmark & 0.168 & 0.561 & 0.071 & 0.737 & 0.959 & 0.993  \\
    & Median      & \xmark & 0.167 & 0.616 & 0.096 & 0.740 & 0.875 & 0.924  \\
    & Median      & \cmark & 0.099 & 0.392 & 0.049 & 0.874 & 0.945 & 0.970  \\
    & Linear Fit  & \xmark & 0.125 & 0.405 & 0.071 & 0.860 & 0.958 & 0.977  \\
    & Linear Fit  & \cmark & \underline{0.033} & \textbf{0.165} & \textbf{0.015} & \textbf{0.984} & \textbf{0.995} & \textbf{0.998}  \\
    & SLF-250   & \cmark & 0.054 & 0.256 & 0.026 & 0.957 & 0.987 & 0.994  \\
    & SLF-500   & \cmark & 0.047 & 0.236 & 0.022 & 0.965 & 0.991 & 0.996  \\
    & SLF-1000  & \cmark & 0.042 & 0.216 & 0.019 & 0.971 & 0.993 & \underline{0.997}  \\
    & SLF-2000  & \cmark & 0.039 & 0.203 & 0.018 & 0.975 & \underline{0.994} & \underline{0.997}  \\
    & SSF-250   & \cmark & 0.046 & 0.234 & 0.022 & 0.963 & 0.989 & 0.995  \\
    & SSF-500   & \cmark & 0.039 & 0.212 & 0.018 & 0.971 & 0.992 & \underline{0.997}  \\
    & SSF-1000  & \cmark & 0.035 & 0.195 & \underline{0.016} & 0.976 & 0.993 & \underline{0.997}  \\
    & SSF-2000  & \cmark & \textbf{0.032} & \underline{0.182} & \textbf{0.015} & \underline{0.979} & \textbf{0.995} & \textbf{0.998}  \\
    \hline 
    \multirow{16}*{Depth Anything v1}
    & Global  & \xmark & 0.199 & 0.646 & 0.087 & 0.630 & 0.926 & 0.987  \\
    & Image   & \xmark & 0.169 & 0.517 & 0.068 & 0.749 & 0.965 & 0.997  \\
    & RSA     & \xmark & 0.147 & 0.484 & 0.065 & 0.775 & 0.975 & 0.997  \\
    & Median       & \xmark & 0.160 & 0.600 & 0.091 & 0.748 & 0.881 & 0.929  \\
    & Median       & \cmark & 0.096 & 0.378 & 0.048 & 0.877 & 0.946 & 0.970  \\
    & Linear Fit   & \xmark & 0.119 & 0.390 & 0.069 & 0.870 & 0.959 & 0.977  \\
    & Linear Fit   & \cmark & \textbf{0.030} & \textbf{0.159} & \textbf{0.014} & \textbf{0.985} & \textbf{0.995} & \underline{0.998}  \\
    & LF-LiDAR 1-beam   & \xmark & 0.063 & 0.652 & 0.028 & 0.939 & 0.981 & 0.993  \\
    & LF-LiDAR 16-beam  & \xmark & 0.039 & 0.454 & 0.017 & 0.976 & \textbf{0.995} & \textbf{0.999}  \\
    & LF-LiDAR 32-beam  & \xmark & 0.040 & 0.461 & 0.017 & 0.974 & \underline{0.994} & \textbf{0.999}  \\
    & SLF-250           & \cmark & 0.050 & 0.249 & 0.024 & 0.959 & 0.989 & 0.995  \\
    & SLF-500           & \cmark & 0.044 & 0.227 & 0.021 & 0.967 & 0.991 & 0.996  \\
    & SLF-1000          & \cmark & 0.039 & 0.209 & 0.018 & 0.973 & 0.993 & 0.997  \\
    & SLF-2000          & \cmark & 0.036 & 0.197 & 0.017 & 0.977 & \underline{0.994} & 0.997  \\
    & SSF-250           & \cmark & 0.044 & 0.229 & 0.021 & 0.964 & 0.990 & 0.996  \\
    & SSF-500           & \cmark & 0.038 & 0.208 & 0.018 & 0.972 & 0.992 & 0.997  \\
    & SSF-1000          & \cmark & 0.034 & 0.190 & \underline{0.016} & 0.977 & \underline{0.994} & 0.997  \\
    & SSF-2000          & \cmark & \underline{0.031} & \underline{0.178} & \textbf{0.014} & \underline{0.980} & \textbf{0.995} & \underline{0.998}  \\
    \hline 
    \multirow{10}*{Depth Anything v2}
    & Median      & \xmark & 0.160 & 0.608 & 0.090 & 0.746 & 0.884 & 0.934  \\
    & Median      & \cmark & 0.092 & 0.370 & 0.045 & 0.883 & 0.951 & 0.973  \\
    & Linear Fit  & \xmark & 0.125 & 0.401 & 0.074 & 0.859 & 0.957 & 0.975  \\
    & Linear Fit  & \cmark & \textbf{0.030} & \textbf{0.161} & \textbf{0.014} & \textbf{0.984} & \textbf{0.995} & \textbf{0.998}  \\
    & SLF-250   & \cmark & 0.051 & 0.250 & 0.025 & 0.957 & 0.987 & 0.994  \\
    & SLF-500   & \cmark & 0.044 & 0.227 & 0.021 & 0.966 & 0.990 & 0.995  \\
    & SLF-1000  & \cmark & 0.039 & 0.210 & 0.018 & 0.972 & 0.992 & 0.996  \\
    & SLF-2000  & \cmark & 0.036 & 0.198 & 0.017 & 0.975 & \underline{0.994} & \underline{0.997}  \\
    & SSF-250   & \cmark & 0.044 & 0.233 & 0.021 & 0.963 & 0.989 & 0.995  \\
    & SSF-500   & \cmark & 0.038 & 0.211 & 0.018 & 0.971 & 0.992 & 0.996  \\
    & SSF-1000  & \cmark & 0.034 & 0.194 & \underline{0.016} & 0.976 & 0.993 & \underline{0.997}  \\
    & SSF-2000  & \cmark & \underline{0.031} & \underline{0.182} & \textbf{0.014} & \underline{0.979} & \underline{0.994} & \underline{0.997}  \\
    \bottomrule
    \end{tabular}
    \vspace{3mm}
    \caption{
    \textbf{Quantitative results on NYU Depth v2 dataset.} 
    Our methods, \textbf{Sparse Linear Fit (SLF)} and \textbf{Sparse Surface Fit (SSF)}, demonstrate strong competitiveness against existing baselines across all evaluation metrics. 
    Global refers to optimizing a single same scale and shift for the entire dataset. 
    Image denotes predicting scales and shifts using CLIP image features. 
    RSA denotes predicting scales and shifts using CLIP text features. 
    Median indicates scaling using the ratio between the median of depth prediction and ground truth. 
    Linear fit denotes optimizing scale and shift to fit to ground truth for each image. 
    DA refers to domain adaptation. 
    ZoeDepth performs per-pixel refinement. 
    LF-LiDAR applies Linear Fit using depth samples from simulated LiDAR beams. 
    For each model, the best result in its category is highlighted in \textbf{bold}, and the second best is \underline{underlined}.
    }
    \vspace*{-5mm}
    \label{tab:nyu_results}
    \end{table*}

    \renewcommand{\arraystretch}{1.1}
    \begin{table*}[ht]
    \centering
    \footnotesize
    \begin{tabular}{l|c|c|ccc|ccc}
    \toprule
    \textbf{Models} & \textbf{Scaling} & \textbf{Region-aware} & Abs Rel $\downarrow$ & $\text{RMSE}_{\log} \downarrow$ & RMSE $\downarrow$ & $\delta<1.25 \uparrow$ & $\delta<1.25^{2} \uparrow$ & $\delta<1.25^{3} \uparrow$  \\  
    \midrule
    \multirow{13}*{MiDas}
    &Global    & \xmark &0.192 &0.212 &4.811 &0.729 &0.939 &0.978\\
    &Image     & \xmark &0.164 &0.199 &4.254 &0.749 &0.949 &0.982\\
    &RSA       & \xmark &0.155 &0.179 &3.989 &0.794 &0.960 &\textbf{0.992}\\
    &Median       & \xmark &0.210 &0.491 &0.393 &0.656 &0.826 &0.893\\
    &Median       & \cmark &0.152 &0.349 &0.315 &0.778 &0.901 &0.946\\
    &Linear fit   & \xmark &0.168 &0.419 &0.622 &0.805 &0.919 &0.953\\
    &Linear fit   & \cmark &0.057 &\textbf{0.128} &0.462 &\underline{0.956} &\textbf{0.984} &\underline{0.991}\\
    & SLF-250    & \cmark & 0.068 & 0.145 & 0.196 & 0.942 & 0.980 & \underline{0.991}  \\
    & SLF-500    & \cmark & 0.063 & 0.137 & 0.185 & 0.948 & 0.982 & \textbf{0.992}  \\
    & SLF-1000   & \cmark & 0.061 & 0.133 & 0.180 & 0.950 & \underline{0.983} & \textbf{0.992}  \\
    & SLF-2000   & \cmark & 0.059 & 0.132 & 0.176 & 0.953 & \underline{0.983} & \textbf{0.992}  \\
    & SSF-250    & \cmark & 0.059 & 0.159 & 0.184 & 0.947 & 0.979 & 0.990  \\
    & SSF-500    & \cmark & 0.056 & 0.147 & 0.177 & 0.951 & 0.981 & \underline{0.991}  \\
    & SSF-1000   & \cmark & \underline{0.053} & 0.139 & \underline{0.167} & 0.955 & \underline{0.983} & \textbf{0.992}  \\
    & SSF-2000   & \cmark & \textbf{0.051} & \underline{0.129} & \textbf{0.160} & \textbf{0.957} & \textbf{0.984} & \textbf{0.992}  \\
    \hline 
    \multirow{13}*{Depth Anything v1}
    &Global    & \xmark & 0.191 & 0.228 &5.273 &0.663 &0.932 &0.981\\
    &Image     & \xmark & 0.162 & 0.195 &4.483 &0.768 &0.951 &0.983\\
    &RSA       & \xmark & 0.147 & 0.179 &4.143 &0.786 &0.967 &\textbf{0.995}\\
    &Median       & \xmark & 0.194 & 0.476 &0.798 &0.690 &0.828 &0.889\\
    &Median       & \cmark & 0.147 & 0.340 &0.634 &0.788 &0.903 &0.945\\
    &Linear fit   & \xmark & 0.160 & 0.421 &0.602 &0.820 &0.921 &0.953\\
    &Linear fit   & \cmark & 0.055 & \textbf{0.124} &0.458 &\textbf{0.958} &\underline{0.984} &0.992\\
    & SLF-250    & \cmark & 0.065 & 0.151 & 0.192 & 0.943 & 0.979 & 0.990  \\
    & SLF-500    & \cmark & 0.062 & 0.140 & 0.184 & 0.947 & 0.981 & 0.990  \\
    & SLF-1000   & \cmark & 0.058 & 0.129 & 0.174 & 0.953 & 0.983 & 0.992  \\
    & SLF-2000   & \cmark & 0.056 & 0.127 & 0.170 & 0.955 & \underline{0.984} & 0.992  \\
    & SSF-250    & \cmark & 0.061 & 0.162 & 0.188 & 0.946 & 0.978 & 0.989  \\
    & SSF-500    & \cmark & 0.056 & 0.144 & 0.176 & 0.951 & 0.981 & 0.991  \\
    & SSF-1000   & \cmark & \underline{0.052} & 0.136 & \underline{0.166} & \underline{0.956} & 0.983 & 0.992  \\
    & SSF-2000   & \cmark & \textbf{0.050} & \underline{0.125} & \underline{0.158} & \textbf{0.958} & \textbf{0.985} & \underline{0.993}  \\
    \hline 
    \multirow{13}*{Depth Anything v2}
    &Median        & \xmark &0.190 &0.461 &0.799 &0.696 &0.836 &0.897\\
    &Median        & \cmark &0.138 &0.325 &0.619 &0.802 &0.913 &0.952\\
    &Linear fit    & \xmark &0.160 &0.428 &0.605 &0.825 &0.923 &0.952\\
    &Linear fit    & \cmark &0.055 &\textbf{0.127} &0.459 &\textbf{0.958} &\underline{0.983} &\underline{0.991}\\
    & SLF-250     & \cmark  & 0.066 & 0.155 & 0.194 & 0.944 & 0.978 & 0.989  \\
    & SLF-500     & \cmark  & 0.062 & 0.144 & 0.188 & 0.947 & 0.980 & 0.990  \\
    & SLF-1000    & \cmark  & 0.059 & 0.137 & 0.180 & 0.951 & 0.982 & \underline{0.991}  \\
    & SLF-2000    & \cmark  & 0.057 & 0.134 & 0.175 & \underline{0.954} & \underline{0.983} & 0.992  \\
    & SSF-250     & \cmark  & 0.061 & 0.170 & 0.191 & 0.945 & 0.978 & 0.989  \\
    & SSF-500     & \cmark  & 0.056 & 0.153 & 0.179 & 0.952 & 0.981 & \underline{0.991}  \\
    & SSF-1000    & \cmark  & \underline{0.053} & 0.143 & \underline{0.171} & \underline{0.954} & 0.982 & \underline{0.991}  \\
    & SSF-2000    & \cmark  & \textbf{0.051} & \underline{0.136} & \textbf{0.163} & \textbf{0.958} & \textbf{0.984} & \textbf{0.992}  \\
    \bottomrule
    \end{tabular}
    \vspace{3mm}
    \caption{\textbf{Quantitative results on VOID dataset.} 
    For each model, the best result in its category is highlighted in \textbf{bold}, and the second best is \underline{underlined}.
     Please refer to Table \ref{tab:nyu_results} for more details about notations.
    }
    \vspace*{-4mm}
    \label{tab:void_results}
    \end{table*}

In our method, we replace the linear regression used in Algorithm 1 with this surface fitting approach, as it aligns better with the underlying physical principles. 
These parameters are computed separately for each region using the surface fitting method, ensuring that the surface fitting is performed in a region-specific manner. Once the fitting is complete for all regions, we merge the results to obtain the final, unified surface. This approach ensures that each region’s depth map is accurately scaled and shifted based on its local characteristics, leading to a more precise and region-aware metric depth map.

\section{Experiments}
\label{sec:experment}

We conducted extensive experiments to evaluate our non-learning-based methods, \textbf{Sparse Linear Fit (SLF)} and \textbf{Sparse Surface Fit (SSF)}. SLF uses {sparse depth measurements} to compute region-aware scale and shift factors to convert relative depth into metric depth and SSF further estimates region-aware scale, coefficient, and shift parameters based on {sparse depth measurements} to achieve the same goal, as detailed in Algorithm~\ref{alg:region_depth_scaling}. We evaluate their performance on standard depth estimation benchmarks, compare them with state-of-the-art (SOTA) methods, and conduct ablation studies to analyze the contribution of each component in our approach.

\subsection{Experimental Setup} 
\textbf{Datasets.}
We present our main experimental results on two datasets: NYUv2~\cite{silberman2012indoor}
and VIOD \cite{wong2020unsupervised}.
The NYUv2 dataset consists of images with a resolution of $480 \times 640$ and depth values ranging from  0.001 to 10 meters. We follow the dataset partitioning method from \cite{lee2019big, liu2023va, zeng2024rsa}, which includes 24,231 training images and 654 test images.
VOID contains images with a resolution of $480 \times 640$ where depth values from 0.2 to 5 meters. It contains 48,248 train images and 800 test images following the official splits \cite{wong2020unsupervised}.
Since our method does not require retraining, we do not use the training sets for either dataset; instead, we only employ their test images for evaluation.
In our experiments, since our method does not require retraining the model, we only used the test split of each dataset and randomly sample 250, 500, 1000, and 2000 depth points from the ground truth data.

\begin{figure*}
  \centering
  \includegraphics[height=112mm]{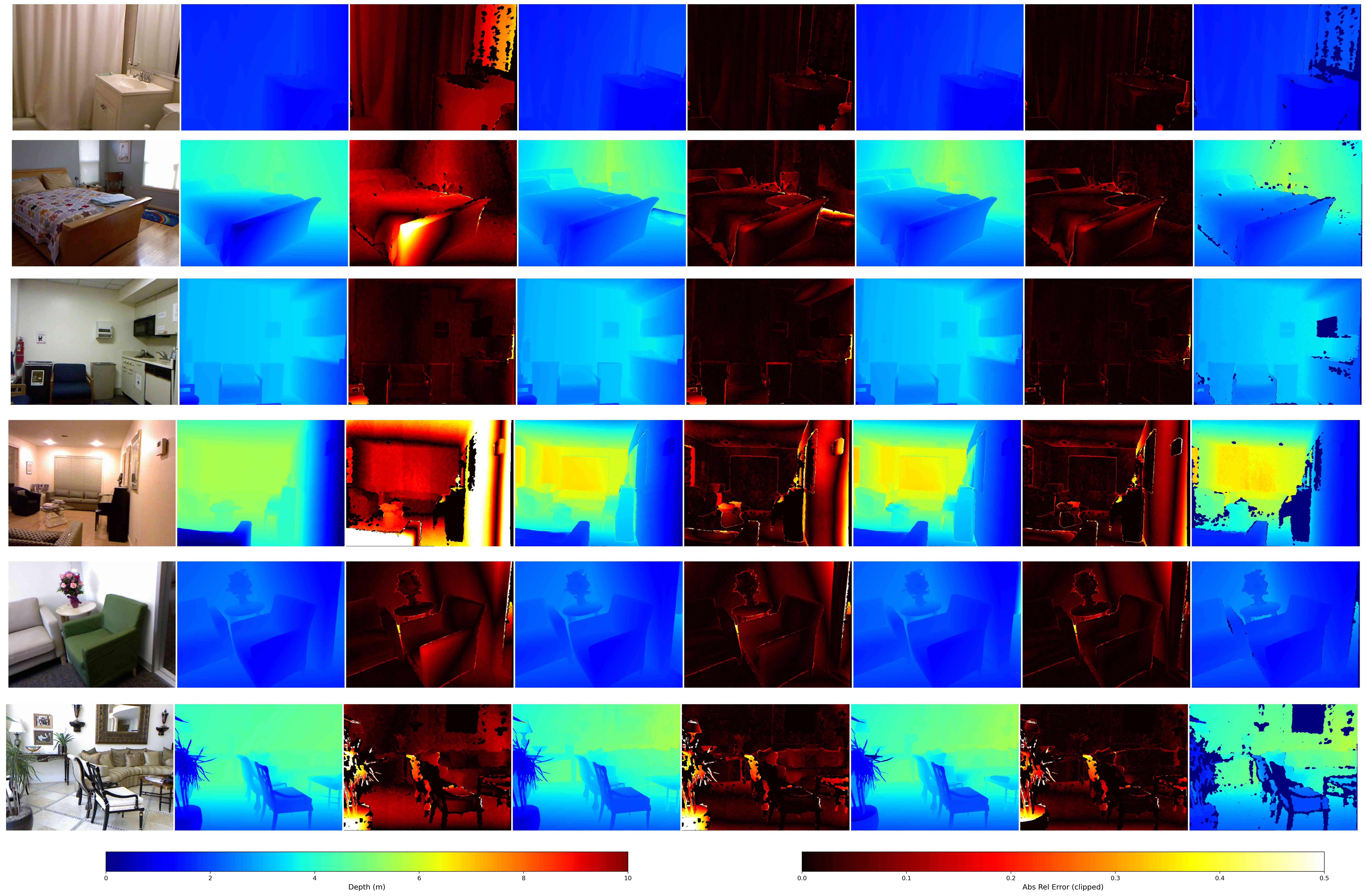}
  \caption{
 Visualization of depth scale adaptation results of Depth Anything V2 on the NYU Depth V2 dataset. From left to right, the images represent: the input image, Linear Fit scaling with its error map, SSF-250 with its error map, SSF-2000 with its error map, and the ground truth. Note: Zeros in the ground truth indicate the absence of valid depth values.
  }
  \label{fig:depth_comparison}
\end{figure*}

\textbf{Foundation models.}
For all the datasets, we adopt Segment Anything (SAM)~\cite{li2025segment} as the image segmentation foundation model due to its zero-shot generalization capability and high precision in handling complex scenes. Trained on 11 million images and over 1 billion masks, SAM demonstrates robust performance even on unseen data distributions, making it particularly suitable for indoor scene segmentation tasks.

For the depth estimation task, we use MiDaS~\cite{ranftl2020towards}, Depth Anything~\cite{depth_anything}, and Depth Anything V2~\cite{depth_anything_v2} as the depth prediction foundation models across all datasets. 
For MiDaS, we use the MiDaS 3.1 Swin2-large-384 model with 213M parameters.
For Depth Anything, we use the Depth-Anything-Large with 335.3M parameters.
For Depth Anything V2, we adopt the Depth-Anything-V2-Large, also with 335.3M parameters.

\textbf{Evaluation metrics. }
We follow the evaluation protocols of previous works~\cite{bhat2021adabins, yuan2022neural, eigen2014depth}, using the following metrics: mean absolute relative error (Abs Rel), root mean square error (RMSE), absolute error in log space ($\log_{10}$), logarithmic RMSE ($\text{RMSE}_{\log}$), and threshold accuracy ($\delta_i$).


\subsection{Quantitative results} 

We present the results on NYU Depth v2 in Table \ref{tab:nyu_results} and VOID in Table \ref{tab:void_results}.
The "Global" scaling, following \cite{ranftl2021vision}, refers to a method where the scale and shift are optimized over the training set and applied to all test samples.
The "Image" scaling, following \cite{zeng2024rsa}, estimates the scale and shift from CLIP image features.
The "RSA" scaling, following \cite{zeng2024rsa}, regresses a linear transformation from CLIP-encoded text captions describing the scene, and applies it globally to the relative depth to produce metric-scaled depth predictions.
The "Linear Fit" scaling, following \cite{ranftl2020towards}, performs linear regression to determine the optimal scale and shift that minimize the least-squares error between predicted and ground truth metric depths.
The "Median" scaling, as in \cite{godard2019digging}, computes the ratio between the median values of predicted and ground truth depths.
The "LF-LiDAR" method, proposed by \cite{marsal2024foundation}, performs a linear regression between the predicted metric depth and LiDAR measurements using 1, 16, and 32-beam LiDAR data.
Our methods, \textbf{SLF} and \textbf{SSF}, perform depth adaptation using 250, 500, 1000, and 2000 depth measurements randomly sampled from the ground truth depth maps.

Compared to ground-truth-based methods such as the non-region-aware Median, Linear Fit, and LF-LiDAR methods, which treat the depth map as a global entity and apply uniform transformations, our methods, \textbf{SLF} and \textbf{SSF}, achieves superior performance while relying on significantly fewer sparse depth measurements.
Median and Linear Fit use the entire ground-truth depth map to compute global scaling factors while LF-LiDAR leverages partial ground truth for scaling. 
For example, in a $640 \times 480$ image, a 1-beam LiDAR provides approximately 640 depth points, while a 32-beam LiDAR yields over 20,000. 
In contrast, our method outperforms them even when using far fewer measurements.
By leveraging the region-level structure of the depth map, our method segments it into semantically and geometrically meaningful regions to enable scale adaptation for relative depth predictions. This region-aware formulation captures local depth characteristics at a finer granularity, allowing for more accurate scale estimation with substantially fewer depth samples. In contrast, non-region-aware methods perform simple global transformations and overlook the inherent regional composition of the scene, often resulting in suboptimal performance. Moreover, when the region-aware formulation is incorporated into the Median and Linear Fit baselines, their performance improves significantly. These findings validate the effectiveness of our region-aware strategy and underscore its potential as a principled approach for improving depth scale adaptation in MDE.
Compared to the Global, Image, and RSA methods, which perform scale adaptation based on the training set, our proposed method achieves significant performance gains even when only a small number of sparse depth measurements (e.g., just 250 points) are introduced.

For both \textbf{SLF} and \textbf{SSF}, as the number of sparse depth measurements increases, their performance improves consistently, indicating that more depth samples result in more accurate fitting.
Moreover, under the same number of depth samples, SSF consistently outperforms SLF, suggesting that surface fitting is more suitable for accurate scale adaptation. This further supports our hypothesis that a depth map is best interpreted as a composition of local planar surfaces, and that directly fitting these surfaces yields more precise metric depth predictions.

In summary, across all evaluation metrics, \textbf{SLF} and \textbf{SSF} demonstrate clear advantages over existing baselines. These results highlight the effectiveness of our sparse fitting strategy, even when compared with methods that utilize ground truth depth for scale adaptation.

\subsection{Qualitative results}

We present the depth scaling results of the Depth Anything v2 model on several scenes from the NYU Depth v2 dataset in Figure~\ref{fig:depth_comparison}. 
Compared with the global scaling method Linear Fit, 
our method maintains more consistent scaling across the image without introducing significant errors. 
Linear Fit, due to its reliance on a single global scaling strategy, often fails in certain structural regions, resulting in large local errors. 
In contrast, our region-aware scaling approach effectively mitigates the limitations of global methods by adapting to local variations. 
When comparing SSF-250 and SSF-2000, we observe a clear and consistent reduction in error, demonstrating that our method enhances depth estimation accuracy across the entire image while preserving the structure and fine details of the depth map. This improvement is evident in the error maps, where darker regions correspond to more accurate scaling and lower errors. 
Moreover, our method significantly preserves the generalization ability of large-scale models like Depth Anything v2, as shown by the post-scaling results. Unlike methods that require retraining, which often compromise model generalization, our approach retains the original model’s robustness while improving metric consistency.

\section{Conclusion}
\label{sec:conclusion}

In this paper, we propose a region-aware depth scale adaptation method for monocular depth estimation foundation models, which can efficiently and accurately recover metric-scale depth without retraining or fine-tuning. The method shows higher accuracy on multiple different scenes and datasets. The core of this method is to segment the scene into regions and assign distinct scaling factors according to sparse depth measurements. This region-level scale adaptation can effectively cope with object-level scale differences and overcome the problem that global scaling factor is prone to failure in complex heterogeneous scenes. Two implementations are proposed in this paper: sparse linear fitting (SLF) and sparse surface fitting (SSF), and extensive experiments are conducted on NYUv2 and VOID datasets. The results show that even when using sparse measurements, our method still significantly outperforms existing foundation models (such as Median, Linear Fit, LF-LiDAR) in terms of accuracy. At the same time, after extending the region-aware formulation into traditional Median and Linear Fit baselines, their performance has been significantly improved, further verifying the versatility and compatibility of our method. 
{
    \small
    \bibliographystyle{ieeenat_fullname}
    \bibliography{main}
}

\end{document}